\pdfoutput=1
\documentclass[letterpaper]{article}
\usepackage{aaai}
\usepackage{times}
\usepackage{helvet}
\usepackage{courier}
\usepackage{fixltx2e}
\usepackage{url}
\usepackage{amsmath}
\usepackage{amsthm}
\usepackage{bbm}
\usepackage{graphicx}
\usepackage{subcaption}
\usepackage{float}
\usepackage[T1]{fontenc}
\newtheorem{defn}{Definition}

\def\etal{\emph{et al.}}
\begin{document}
%
\title{Training Sparse Neural Networks}
\author{Suraj Srinivas\\
  Indian Institute of Science\\
  Bangalore, India\\
  \texttt{\small surajsrinivas@grads.cds.iisc.ac.in}
  \And
  Akshayvarun Subramanya\\
  Indian Institute of Science\\
  Bangalore, India\\
  \texttt{\small akshayvarun07@gmail.com}\\
  \And
  R. Venkatesh Babu\\
  Indian Institute of Science\\
  Bangalore, India\\
  \texttt{\small venky@cds.iisc.ac.in} 
}

\maketitle
\begin{abstract}
\begin{quote}
Deep neural networks with lots of parameters are typically used for large scale computer vision tasks such as image classification. This is a result of using dense matrix multiplications and convolutions. However, sparse computations are known to be much more efficient. In this work, we train and build neural networks which implicitly use sparse computations. We introduce additional gate variables to perform parameter selection and show that this is equivalent to using a spike-and-slab prior. We experimentally validate our method on both small and large networks and achieve state-of-the-art compression results for sparse neural network models.
\end{quote}
\end{abstract}
\section{Introduction}

For large-scale tasks such as image classification, large networks with many millions of parameters are often used \cite{krizhevsky2012imagenet}, \cite{Simonyan15}, \cite{Szegedy_2015_CVPR}. However, these networks typically use dense computations. Would it be advantageous to use sparse computations instead? Apart from having fewer number of parameters to store ($\mathcal{O}(mn)$ to $\mathcal{O}(k)$)\footnote{For a matrix of size $m \times n$ with $k$ non-zero elements}, sparse computations also decrease feedforward evaluation time ($\mathcal{O}(mnp)$ to $\mathcal{O}(kp)$)\footnote{For matrix-vector multiplies with a dense vector of size $p$}. Further, having a lower parameter count may help in avoiding overfitting. 

Regularizers are often used to discourage overfitting. These usually restrict the magnitude ($\ell_2 / \ell_1$) of weights. However, to restrict the computational complexity of neural networks, we need a regularizer which restricts the total number of parameters of a network. A common strategy to obtain sparse parameters is to apply sparsity-inducing regularizers such as the $\ell_1$ penalty on the parameter vector. However, this is often insufficient  in inducing sparsity in case of large non-convex problems like deep neural network training as shown in \cite{DBLP:journals/corr/CollinsK14}. The contribution of this paper is to be able to induce sparsity in a tractable way for such models.

The overall contributions of the paper are as follows.
\begin{itemize}
\item We propose a novel regularizer that restricts the total number of parameters in the network. (Section 2)
\item We perform experimental analysis to understand the behaviour of our method. (Section 4)
\item We apply our method on LeNet-5, AlexNet and VGG-16 network architectures to achieve state-of-the-art results on network compression. (Section 4)
\end{itemize}

\section{Problem Formulation}
To understand the motivation behind our method, let us first define our notion of computational complexity of a neural network.

Let $\Phi = \{g^s_1,g^s_2,...,g^s_m\}$ be a set of $m$ vectors. This represents an $m$-layer dense neural network architecture where $g^s_i$ is a vector of parameter indices for the $i^{th}$ layer, i.e;  $g^s_i = \{0,1\}^{n_i}$. Here, each layer $g^s_i$ contains $n_i$ elements. Zero indicates absence of a parameter and one indicates presence. Thus, for a dense neural network, $g_i$ is a vector of all ones, i.e.; $g^s_i = \{1\}^{n_i}$. For a sparse parameter vector, $g^s_i$ would consist of mostly zeros. Let us call $\Phi$ as the index set of a neural network.

For these vectors, our notion of complexity is simply the total number of parameters in the network. 

\begin{defn}
The complexity of a $m$-layer neural network with index set $\Phi$ is given by $\| \Phi \| = \sum\limits_{i=1}^{m} n_i $. 
\label{defn:complexity}
\end{defn}

We now aim to solve the following optimization problem.

\begin{equation}
\hat{\theta}, \hat{\Phi} = \underset{\theta,\Phi}{\arg\min} ~\ell(\hat{y}(\theta, \Phi),y) + \lambda \| \Phi \| 
\label{eqn:main}
\end{equation}

where $\theta$ denotes the weights of the neural network, and $\Phi$ the index set. $\ell(\hat{y}(\theta,\Phi),y)$ denotes the loss function, which depends on the underlying task to be solved. Here, we learn both the weights as well as the index set of the neural network. Using the formalism of the index set, we are able to penalize the total number of network parameters. While easy to state, we note that this problem is difficult to solve, primarily because $\Phi$ contains elements $\in \{0,1\}$. 

\begin{figure*}[h]
\centering
\includegraphics[width=17cm]{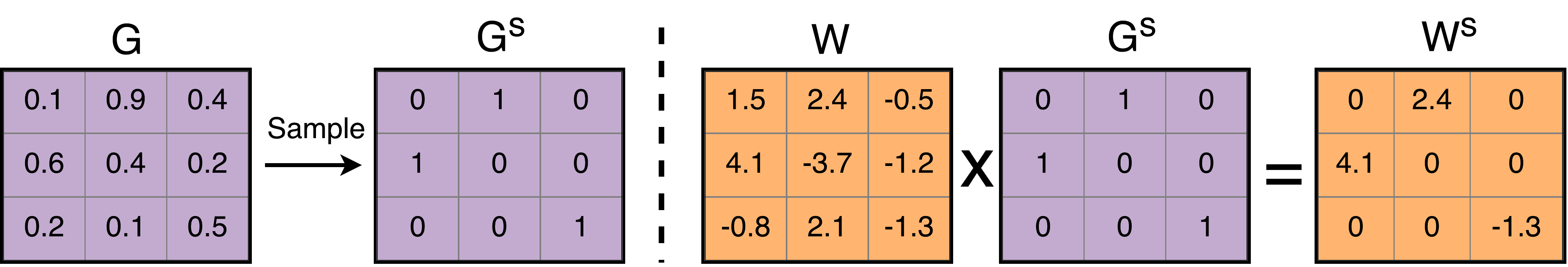}
\caption{\small{Our strategy for sparsifying weight matrices. First, we sample / threshold the gate variables. We then multiply the resulting binary matrix with $W$, to yield a sparse matrix $W^s$.}}
\label{fig:BlockDiag}
\end{figure*}

\subsection{Gate Variables}
How do we incorporate the index set formalism in neural networks? Assume that the index set ($G^s$ in Fig. \ref{fig:BlockDiag}) is multiplied pointwise with the weight matrix. This results in a weight matrix that is \textit{effectively} sparse, if the index set has lots of zeros rather than ones. In other words, we end up learning two sets of variables to ensure that one of them - weights - becomes sparse. How do we learn such binary parameters in the first place ?

To facilitate this, we interpret index set variables ($G^s$) as draws from a bernoulli random variable. As a result, we end up learning the real-valued bernoulli parameters ($G$ in Fig. \ref{fig:BlockDiag}), or \textit{gate variables} rather than index set variables themselves. Here the sampled binary gate matrix $G^s$ corresponds exactly to the index set, or the $\Phi$ matrix described above. To clarify our notation, $G$ and $g$ stand for the real-valued gate variables, while the superscript $(.)^s$ indicates binary sampled variables.

When we draw from a bernoulli distribution, we have two choices - we can either perform a \textit{unbiased} draw (the usual sampling process), or we can perform a so-called \textit{maximum-likelihood (ML)} draw. The ML draw involves simply thresholding the values of $G$ at $0.5$. To ensure determinism, we use the ML draw or thresholding in this work. 

\subsection{Promoting Sparsity}
Given our formalism of gate variables, how do we ensure that the learnt bernoulli parameters are low -  or in our case - mostly less than $0.5$ ? One plausible option is to use the $\ell_2$ or the $\ell_1$ regularizer on the gate variables. However, this does not ensure that there will exist values greater than $0.5$. To accommodate this, we require a \textit{bi-modal} regularizer, i.e; a regularizer which ensures that some values are large, but most values are small.

\begin{figure*}[!t]
\begin{subfigure}{.48\textwidth}
  \centering
\includegraphics[width=6cm]{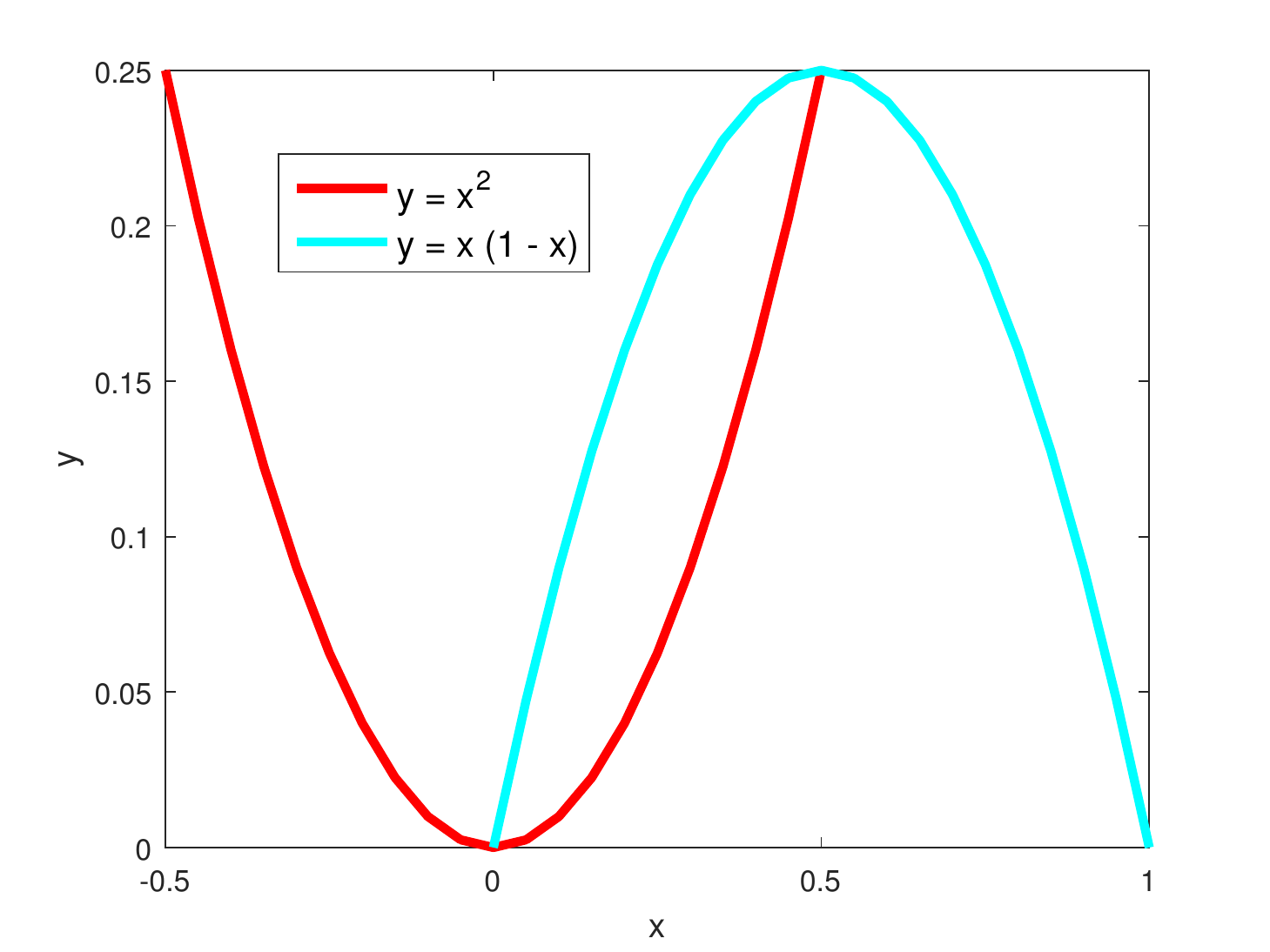}
\caption{{Bi-modal regularizer}}
\label{fig:nngates}
\end{subfigure} %
\begin{subfigure}{.5\textwidth}
  \centering
\includegraphics[width=6cm]{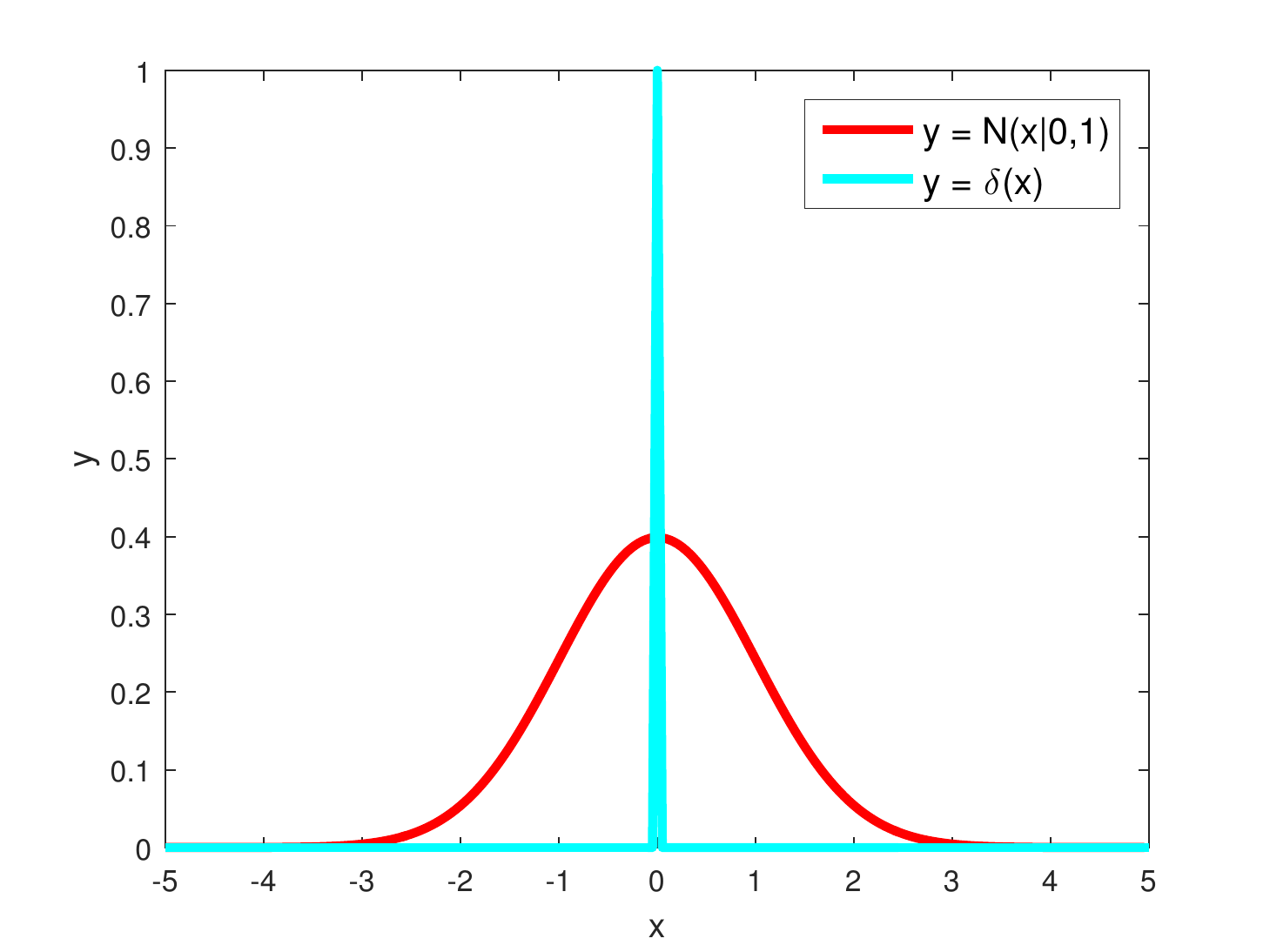}
\caption{{Spike and slab prior}}
\label{fig:betadist}
\end{subfigure}%
\caption{{\textbf{(a)} The bi-modal regularizer used in our work. Note that this encourages values to be close to $0$ and $1$, in contrast to the $\ell_2$ regularizer.
\textbf{(b)} An example of a spike-and-slab prior similar to the one used in this work (except for a constant).}}
\end{figure*}

To this end, we use a regularizer given by $w \times (1-w)$. This was introduced by \cite{murray2010algorithm} to learn binary values for parameters. However, what is important for us is that this regularizer has the \textit{bi-modal} property mentioned earlier, as shown in Fig. \ref{fig:nngates}

Our overall regularizer is simply a combination of this \textit{bi-modal} regularizer as well the traditional $\ell_2$ or $\ell_1$ regularizer for the individual gate variables. Our objective function is now stated as follows.

\begin{equation}
\begin{split}
\hat{\theta}, \hat{\Phi} = \underset{\theta,\Phi}{\arg\min} ~\ell(\hat{y}(\theta, \Phi),y) + \lambda_1 \sum \limits_{i=1}^{m} \sum\limits_{j=1}^{n_i} g_{i,j} (1 - g_{i,j}) \\+ \lambda_2 \sum\limits_{i=1}^{m} \sum\limits_{j=1}^{n_i} g_{i,j} 
\end{split}
\label{eqn:actual}
\end{equation}

where $g_{i,j}$ denotes the $j^{th}$ gate parameter in the $i^{th}$ layer. Note that for $g_{i,j} \in \{0,1\}$, the second term in Eqn. \ref{eqn:actual} vanishes and the third term becomes $\lambda \|\Phi\|$, thus reducing to Eqn.\ref{eqn:main}.

\subsection{An Alternate Interpretation}
Now that we have arrived at the objective function in Eqn.\ref{eqn:actual}, it is natural to ask the question - how do we know that it solves the original objective in Eqn.\ref{eqn:main} ? We shall now derive Eqn.\ref{eqn:actual} from this perspective.

Assuming the formulation of gate variables, we can re-write the objective in Eqn.\ref{eqn:main} as follows.

\begin{eqnarray}
\hat{\theta}, \hat{\Phi} &=& \underset{\theta,G}{\arg\min} ~\ell(\hat{y}(\theta, G^s),y) + \lambda \sum\limits_{i=1}^{m} \sum\limits_{j=1}^{n_i} g^s_{i,j}  \label{eqn:main1} \\
g^s_{i,j} & \sim & \mathrm{bernoulli}(g_{i,j}), ~\forall~ i,j \nonumber
\end{eqnarray}

where $g^s$ is the sampled version of gate variables $g$. Note that Eqn.\ref{eqn:main1} is a stochastic objective function, arising from the fact that $g^s$ is a random variable. We can convert this to a real-valued objective by taking expectations. Note that expectation of the loss function is difficult to compute. As a result, we approximate it with a Monte-Carlo average.

\begin{eqnarray} 
\hat{\theta}, \hat{\Phi}  &=&  \underset{\theta,G}{\arg\min} ~ \frac{1}{t} \sum_t (\ell(\hat{y}(\theta, G^s),y)) + \lambda \sum\limits_{i=1}^{m} \sum\limits_{j=1}^{n_i} g_{i,j} \nonumber  \\
g^s_{i,j}  &\sim& \mathrm{bernoulli}(g_{i,j}), ~ \forall ~i,j  \nonumber
\end{eqnarray}

where $\mathbf{E}(g^s_{i,j}) = g_{i,j}$. While this formulation is sufficient to solve the original problem, we impose another condition on this objective. We would like to minimize the number of Monte-Carlo evaluations in the loss term. This amounts to reducing $[\frac{1}{t} \sum_t (\ell(\hat{y}(\theta, G^s),y)) - \mathbf{E} (\ell(\hat{y}(\theta, G^s),y))]^2$ for a fixed $t$, or reducing the variance of the loss term. This is done by reducing the variance of $g^s$, the only random variable in the equation. To account for this, we add another penalty term corresponding to $\mathbf{Var}(g^s) = g \times (1 - g)$. Imposing this additional penalty and then using $t = 1$ gives us back Eqn.\ref{eqn:actual}.

\subsection{Relation to Spike-and-Slab priors}
We observe that our problem formulation closely resembles spike-and-slab type priors used in Bayesian statistics for variable selection \cite{mitchell1988bayesian}. Broadly speaking, these priors are mixtures of two distributions - one with very low variance (spike), and another with comparatively large variance (slab). By placing a large mass on the spike, we can expect to obtain parameter vectors with large sparsity.

Let us consider for a moment using the following prior for weight matrices of neural networks.

\begin{equation}
P( W ) = \frac{1}{Z} \prod_i  ~ exp(-~(1 - \delta (w_i)))^{\alpha} ~ \mathcal{N}(w_i | 0, \sigma^2)^{1 - \alpha} 
\label{eqn:scalemixture}
\end{equation}

Here, $\delta (\cdot)$ denotes the dirac delta distribution, and $Z$ denotes the normalizing constant, and $\alpha$ is the mixture coefficient. Also note that like \cite{mitchell1988bayesian}, we assume that $w_i \in [-k,k]$ for some $k > 0$. This is visualized in Fig. \ref{fig:betadist}. Note that this is a multiplicative mixture of distributions, rather than additive. By taking negative logarithm of this term and ignoring constant terms, we obtain

\begin{equation}
- \log P( W ) =   - \alpha \sum_i (1 - \delta (w_i)) ~ + ~ \frac{1 - \alpha}{2 \sigma^2} \sum_i w_i^2   
\label{eqn:scalemixturesimplfied}
\end{equation}

Note that the first term in this expression corresponds exactly to the number of non-zero parameters, i.e; the $ \lambda ~ \| \Phi \|$ term of Eqn. \ref{eqn:main}. The second term corresponds to the usual $\ell_2$ regularizer on the weights of the network (rather than gates). As a result, we conclude that Eqn. \ref{eqn:scalemixture} is a spike-and-slab prior which we implicitly end up using in this method.

\subsection{Estimating gradients for gate variables}
How do we estimate gradients for gate variables, given that they are binary stochastic variables, rather than real-valued and smooth? In other words, how do we backpropagate through the bernoulli sampling step? Bengio \emph{et al.} \shortcite{bengio2013estimating} investigated this problem and empirically verified the efficacy of different possible solutions. They conclude that the simplest way of computing gradients - the \textit{straight-through} estimator works best overall. Our experiments also agree with this observation.

The \textit{straight-through} estimator simply involves back-propagating through a stochastic neuron as if it were an identity function. If the sampling step is given by $g^s \sim bernoulli(g)$, then the gradient $\frac{dg^s}{dg} = 1$ is used.

Another issue of consideration is that of ensuring that $g$ always lies in $[0,1]$ so that it is a valid bernoulli parameter. Bengio \emph{et al.} \shortcite{bengio2013estimating} use a sigmoid activation function to achieve this. Our experiments showed that clipping functions worked better. This can be thought of as a `linearized' sigmoid. The clipping function is given by the following expression.

\begin{equation*}
clip(x) = \begin{cases}
1, & x \geq 1 \\
0, & x \leq 0 \\
x, & otherwise
\end{cases}
\end{equation*}

The overall sampling function is hence given by $g^s \sim bernoulli(clip(g))$, and the straight-through estimator is used to estimate gradients overall.

\subsection{Comparison with LASSO}
LASSO is commonly used method to attain sparsity and perform variable selection. The main difference between the above method and LASSO is that  LASSO is primarily a shrinkage operator, i.e.; it shrinks all parameters until lots of them are close to zero. This is not true for the case of spike-and-slab priors, which can have high sparsity and encourage large values at the same time. This is due to the richer parameterization of these priors.

\begin{figure*}[h]
\begin{subfigure}{.33\textwidth}
  \centering
\includegraphics[width=6cm]{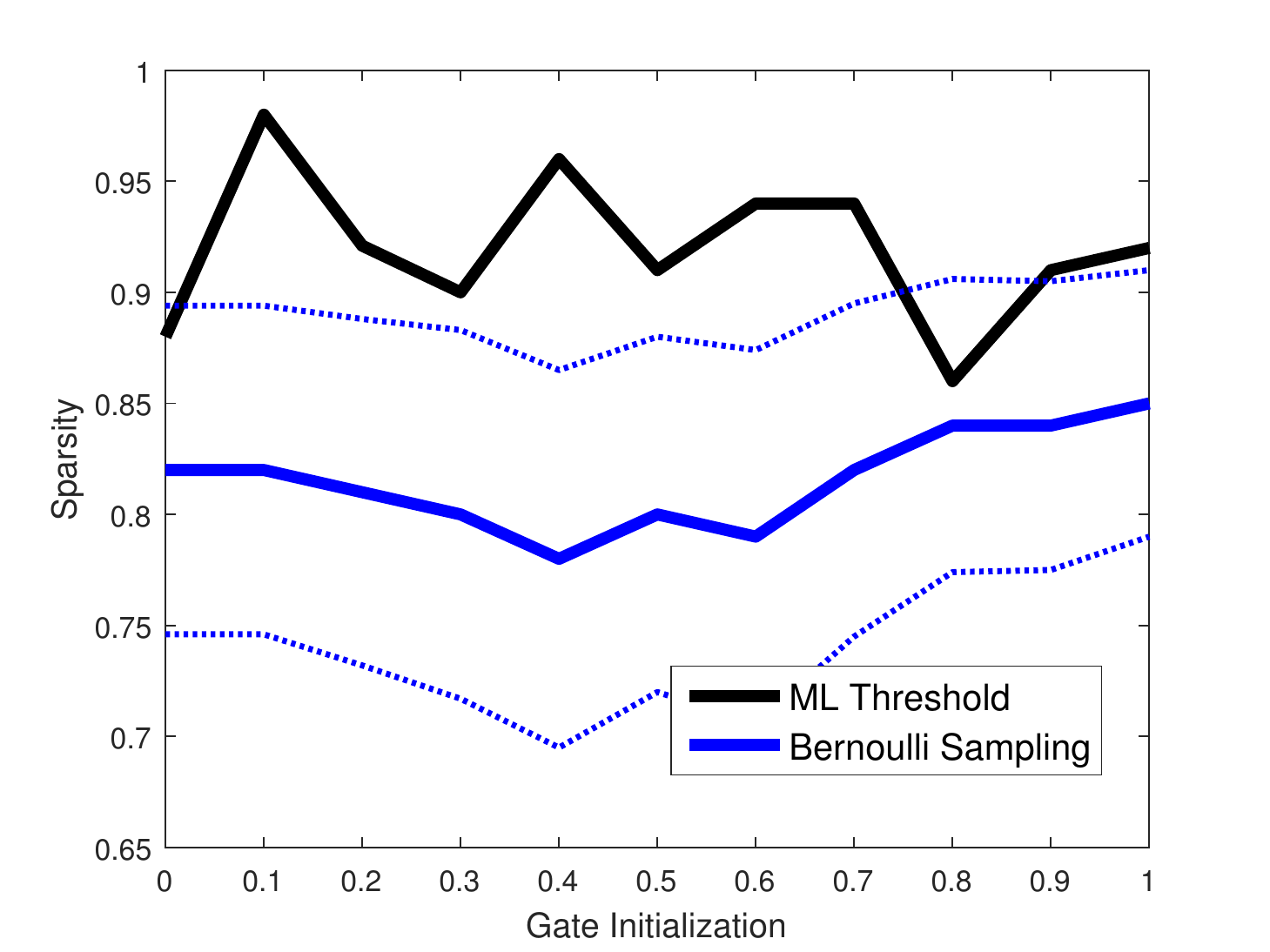}
\caption{{Effect of Gate initialization}}
\label{fig:analyzeinit}
\end{subfigure} %
\begin{subfigure}{.33\textwidth}
  \centering
\includegraphics[width=6cm]{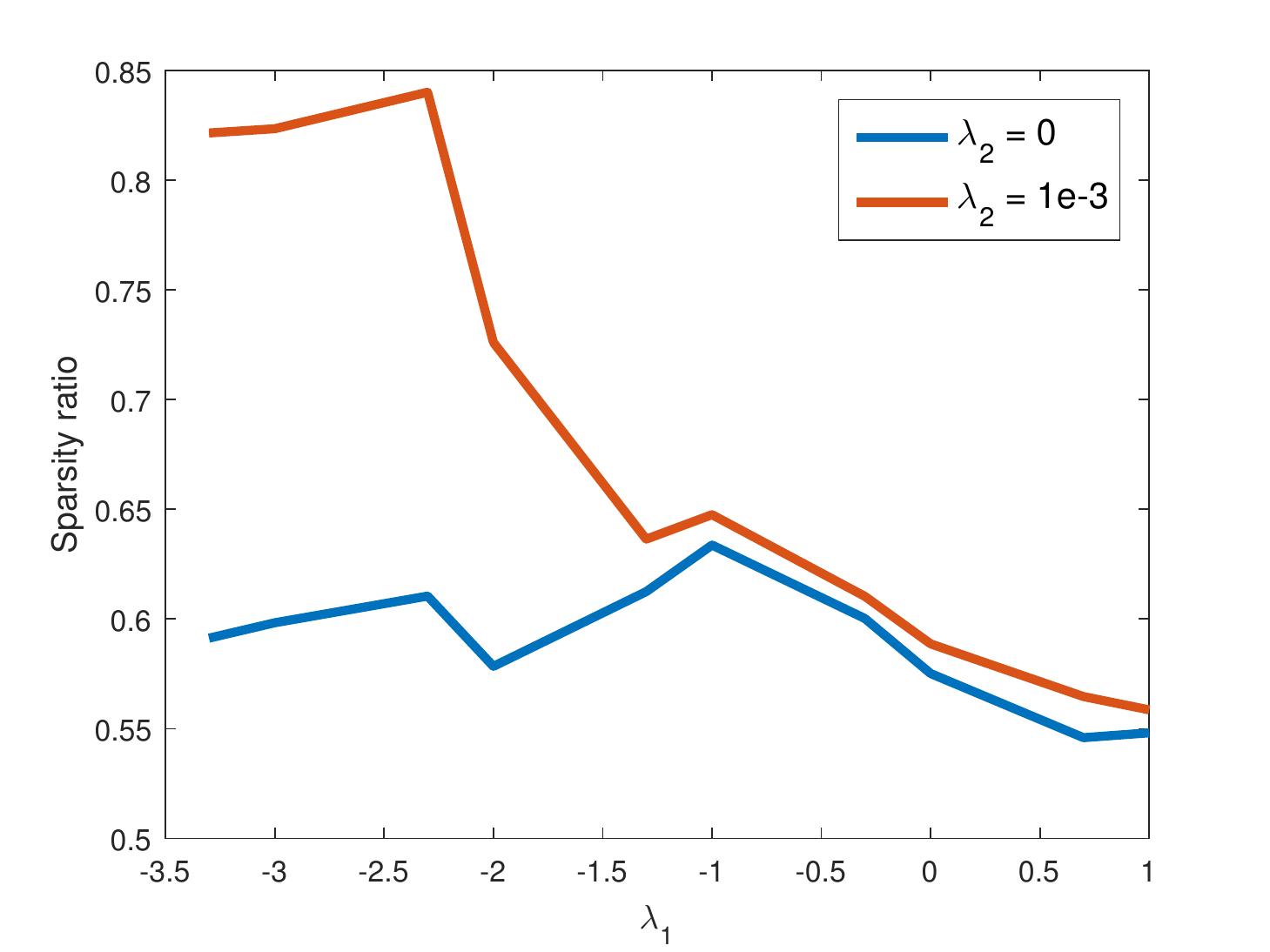}
\caption{{Effect of varying $\lambda_1$}}
\label{fig:lambda1}
\end{subfigure} %
\begin{subfigure}{.33\textwidth}
  \centering
\includegraphics[width=6cm]{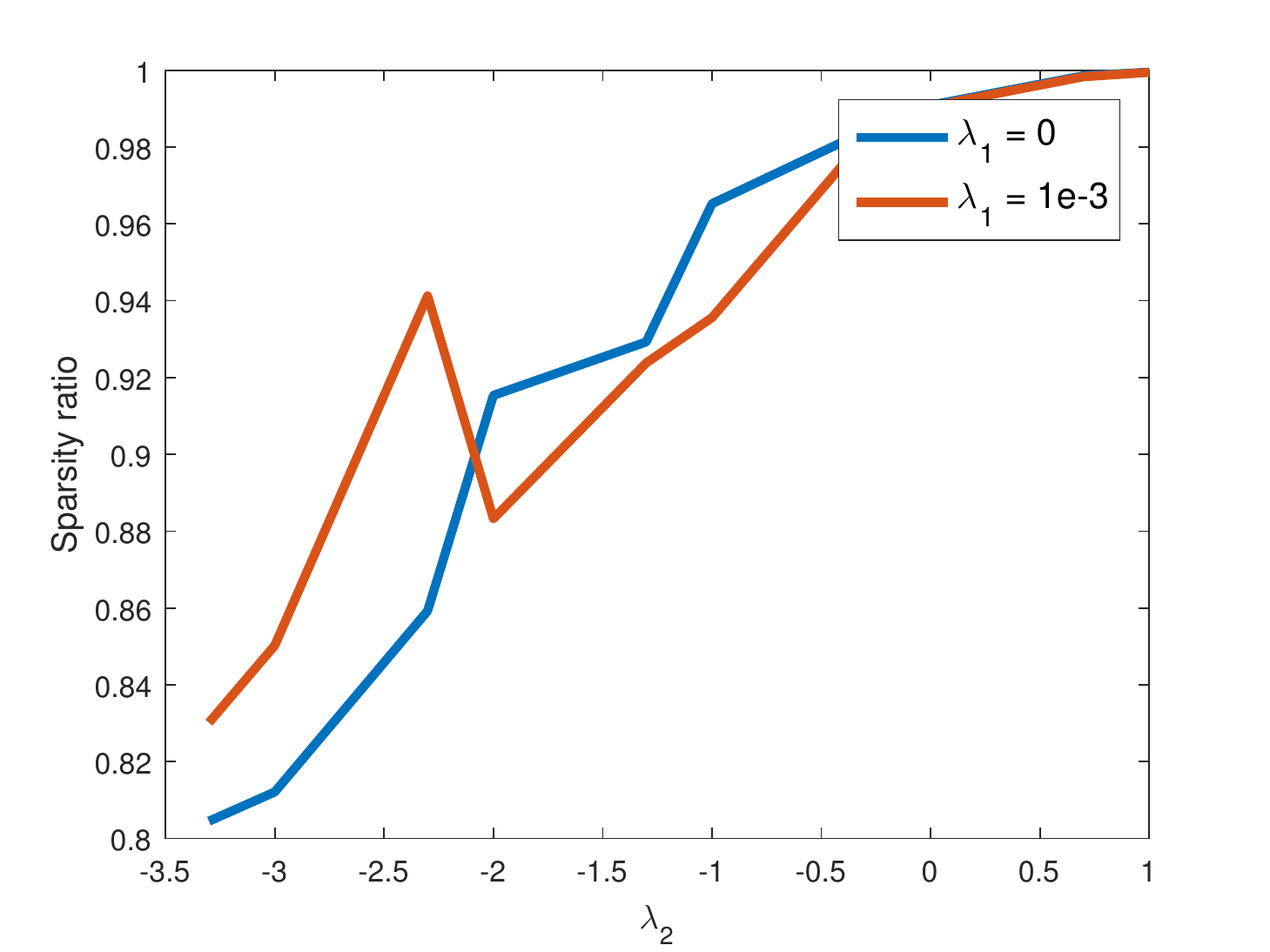}
\caption{{Effect of varying $\lambda_2$}}
\label{fig:lambda2}
\end{subfigure} %
\caption{{\textbf{(a)} We vary the initialization of the gate variables and observe it's effect on sparsity. The dotted blue lines denote the variance of sparsity in case of the sampling-based method.
\textbf{(b)} $\lambda_1$ seems to have a \textit{stabilizing} effect on sparsity whereas \textbf{(c)} increasing $\lambda_2$ seems to increase sparsity.}}
\end{figure*}

\subsection{Practical issues}
In this section we shall discuss some practical issues pertaining to our method. Our method ironically uses twice the number of parameters as a typical neural network, as we have two sets of variables - weights and gates. As a result, model size doubles while training. However, we multiply the two to result in sparse matrices which considerably reduces model size at test time. Essentially we do not have to store both sets of parameters while testing,only a element-wise product of the two is required. Even though the model size doubles at train time, we note that speed of training / feedforward evaluation is not affected due to the fact that only element-wise operations are used.

Our method can be applied to both convolutional tensors as well as fully connected matrices. However while performing compression, we note that convolutional layers are less susceptible to compression that fully connected layers due to the small number of parameters they possess.

\section{Related Work}
There have been many recent works which perform compression of neural networks. Weight-pruning techniques were popularized by LeCun \etal \shortcite{lecun1989optimal} and Hassibi \etal \shortcite{hassibi1993second}, who introduced \emph{Optimal Brain Damage} and \emph{Optimal Brain Surgery} respectively. Recently, Srinivas and Babu \shortcite{BMVC2015_31} proposed a neuron pruning technique, which relied on neuronal similarity. In contrast, we perform weight pruning based on learning, rather than hand-crafted rules.

Previous attempts have also been made to sparsify neural networks. Han \etal \shortcite{han2015learning} create sparse networks by alternating between weight pruning and network training. A similar strategy is followed by Collins and Kohli \shortcite{DBLP:journals/corr/CollinsK14}. On the other hand, our method performs both weight pruning and network training \textbf{simultaneously}. Further, our method has considerably less number of hyper-parameters to determine ($\lambda_1 , \lambda_2$) compared to the other methods, which have $n$ thresholds to be set for each of the $n$ layers in a neural network.


Many methods have been proposed to train models that are deep, yet have a lower parameterisation than conventional networks. Denil \etal \shortcite{denil2013predicting} demonstrated that most of the parameters of a model can be \textit{predicted} given only a few parameters. At training time, they learn only a few parameters and predict the rest. Yang \etal \shortcite{yang2014deep} propose an \emph{Adaptive Fastfood transform}, which is an efficient re-parametrization of fully-connected layer weights. This results in a reduction of complexity for weight storage and computation. Novikov \etal \shortcite{novikov2015tensorizing} use tensor decompositions to obtain a factorization of tensors with small number of parameters. Cheng \etal \shortcite{cheng2015exploration} make use of circulant matrices to re-paramaterize fully connected layers. Some recent works have also focussed on using approximations of weight matrices to perform compression. Gong \etal \shortcite{gong2014compressing} use a clustering-based product quantization approach to build an indexing scheme that reduces the space occupied by the matrix on disk. Note that to take full advantage of these methods, one needs to have fast implementations of the specific parameterization used. One the other hand, we use a sparse parameterization, fast implementations of which are available on almost every platform.

Our work is very similar to that of Architecture Learning \cite{srinivas2015learning}, which uses a similar framework to minimize the total number of neurons in a neural network. On the other hand, we minimize the total number of weights.
\begin{table}[!h]
\centering
\begin{tabular}{ |c|c|c|c|}
 \hline
 \textbf{Layers} & \multicolumn{1}{p{1.5cm}|}{\centering \textbf{Initial } \\ \textbf{Params}} & \multicolumn{1}{p{1.5cm}|}{\centering \textbf{ Final } \\ \textbf{Params}} & \textbf{Sparsity (\%)}\\
 \hline
 \hline
 conv1  & 0.5K    &0.04K & 91\\
 conv2&   25K  & 1.78K &  92.8\\
 fc1 & 400K & 15.4K & 96.1\\
 fc2  & 5K & 0.6K & 86.8\\
 \hline
Total & 431K & 17.9K  & 95.84  \\
\hline
\end{tabular}

   \caption{Compression results for LeNet-5 architecture.}
    \label{table:lenetdetailed}
\end{table}

\begin{table*}[t]
\centering
\begin{tabular}{ |c|c|c|c| }
 \hline
 \textbf{Method} & \textbf{Params (P+I)*} & \textbf{Accuracy(\%)} & \textbf{Compression Rate(\%)} \\
 \hline
 \hline
 Baseline  & 431K    &  99.20 & 1x\\
 SVD(rank-10)\cite{denton2014exploiting}& 43.6K & 98.47 & 10x \\
 AL \cite{srinivas2015learning}  & 40.9K & 99.04 & 10.5x \\
 AF-1024 \cite{yang2014deep} & 38.8K & 99.29 & 11x \\ 
 Han \etal \cite{han2015learning} & 36K (72k)*   & 99.23 & 12x \\
 \hline
 \textbf{Method-1}  & 18K (36k)* & 99.19 & 24x \\
 \textbf{Method-2}  & 22K (44k)* & 99.33 & 19x \\

\hline
\end{tabular}

   \caption{Comparison of compression performance on LeNet-5 architecture. *Count of number of parameters and the indices to store them. This is the effective storage requirement when using sparse computations.}
    \label{table:lenetcompare}
\end{table*}

\section{Experiments}
In this section we perform experiments to evaluate the effectiveness of our method. First, we perform some experiments designed to understand typical behaviour of the method. These experiments are done primarily on LeNet-5 \cite{lecun1998gradient}. Second, we use our method to perform network compression on two networks - LeNet-5 and AlexNet. These networks are trained on MNIST and ILSVRC-2012 dataset respectively. Our implementation is based on Lasagne, a Theano-based library.

\subsection{Analysis of Proposed method}
We shall now describe experiments to analyze the behaviour of our method. First, we shall analyze the effect of hyper-parameters. Second, we study the effect of varying model sizes on the resulting sparsity. 

For all analysis experiments, we consider the LeNet-5 network. LeNet-5 consists of two $5 \times 5$ convolutional layers with 20 and 50 filters, and two fully connected layers with 500 and 10 (output layer) neurons. For analysis, we only study the effects sparsifying the third fully connected layer.
\begin{table}[h!]
\centering
\begin{tabular}{ |c|c|c|c|}
 \hline
 \textbf{Layers} & \multicolumn{1}{p{1.4cm}|}{\centering \textbf{Initial } \\ \textbf{Params}} & \multicolumn{1}{p{1.4cm}|}{\centering \textbf{Final} \\ \textbf{Params}} & \textbf{Sparsity (\%)}\\
 \hline
 \hline
 conv(5 layers)  & 2.3M  & 2.3M & -\\
 fc6 & 38M & 1.3M & 96.5 \\
 fc7  & 17M & 1M & 94\\
 fc8  & 4M & 1.2M & 70\\
 \hline
Total & 60.9M & 5.9M  & 90 \\
\hline
\end{tabular}

   \caption{Layer-wise compression performance on AlexNet}
    \label{table:4}
\end{table}

\begin{table}
\centering
\begin{tabular}{ |c|c|c|c|}
 \hline
 \textbf{Layers} & \multicolumn{1}{p{1.4cm}|}{\centering \textbf{Initial } \\ \textbf{Params}} & \multicolumn{1}{p{1.4cm}|}{\centering \textbf{Final} \\ \textbf{Params}} &  \textbf{Sparsity(\%)}\\
 \hline
 \hline
 \multicolumn{1}{| p{2cm}|}{\centering conv1\textunderscore1 to conv4\textunderscore3 \\ (10 layers)}  & 6.7M  & 6.7M & -\\
 conv5\textunderscore1 & 2M & 2M & - \\
 conv5\textunderscore2 & 2M & 235K & 88.2\\
 conv5\textunderscore3 & 2M & 235K & 88.2\\
 fc6 & 103M & 102K & 99.9  \\
 fc7  & 17M & 167K & 99.01 \\
 fc8  & 4M & 409K & 89.7 \\
 \hline
Total & 138M & 9.85M  & 92.85 \\
\hline
\end{tabular}

   \caption{Layer-wise compression performance on VGG-16}
    \label{table:5}
\end{table}
\subsubsection{Effect of hyper-parameters}
In Section 2.1 we described that we used maximum likelihood sampling (i.e.; thresholding) instead of unbiased sampling from a bernoulli. In these experiments, we shall study the relative effects of hyper-parameters on both methods. In the sampling case, sparsity is difficult to measure as different samples may lead to slightly different sparsities. As a result, we measure expected sparsity as well the it's variance. 

Our methods primarily have the following hyper-parameters: $\lambda_1, \lambda_2$ and the initialization for each gate value. As a result, if we have a network with $n$ layers, we have $n+2$ hyper-parameters to determine. 

First, we analyze the effects of $\lambda_1$ and $\lambda_2$. We use different combinations of initializations for both and look at it's effects on accuracy and sparsity. As shown in Table \ref{table:analyzelambda}, both the thresholding as well as the sparsity-based methods are similarly sensitive to the regularization constants.

In Section 2.3, we saw that $\lambda_1$ roughly controls the variance of the bernoulli variables while $\lambda_2$ penalizes the mean. In Table \ref{table:analyzelambda}, we see that the mean sparsity for the pair $(\lambda_1, \lambda_2) = (0,1)$ is high, while that for $(1,0)$ is considerably lower. Also, we note that the variance of $(1,1)$ is smaller than that of $(0,1)$, confirming our hypothesis that $\lambda_1$ controls variance. 

Overall, we find that both networks are almost equally sparse, and that they yield very similar accuracies. However, the thresholding-based method is deterministic, which is why we primarily use this method.

To further analyze effects of $\lambda_1$ and $\lambda_2$, we plot sparsity values attained by our method by fixing one parameter and varying another. In Figure \ref{fig:lambda1} we see that $\lambda_1$, or the variance-controlling hyper-parameter, mainly \textit{stabilizes} the training by reducing the sparsity levels. In Figure \ref{fig:lambda2} we see that increasing $\lambda_2$ increases the sparsity level as expected.

\begin{table}[!h]
\centering
\begin{tabular}{|c|c||c||c|c|}
 \hline
 $\lambda_1$ & $\lambda_2$ & \multicolumn{1}{p{1.5cm}||}{\centering \textbf{Sparsity (\%)} \\ \textbf{[T]}}  & \multicolumn{1}{p{2cm}|}{\centering \textbf{Avg.Sparsity (\%)} \\ \textbf{[S]}}  & \multicolumn{1}{p{1.5cm}|}{\centering \textbf{Variance (\%)} \\ \textbf{[S]}} \\
 \hline
 0 & 0  & 54.5 & 53.1 & 16.1 \\
 1  & 1   & 98.3 & 93.7 & 3.3\\
 1  & 0    & 62.1 & 57.3 & 5.4\\
 0  & 1    & 99.0 & 92.7 & 4.1 \\
 \hline
\end{tabular}
  \caption{Effect of $\lambda$ parameters on sparsity. [T] denotes the threshold-based method, while [S] denotes that sampling-based method.}
 \label{table:analyzelambda}
\end{table}

\begin{table*}[!th]
\centering
\begin{tabular}{ |c|c|c |c| }
 \hline
 \textbf{Method} & \textbf{Params (Params + Indices)} & \textbf{Top-1 Accuracy(\%)} & \textbf{Compression Rate}\\
 \hline
 \hline
 Baseline  & 60.9M   &  57.2 & - \\
 Neuron Pruning \cite{BMVC2015_31} &   39.6M & 55.60 & 1.5x \\
 SVD-quarter-F \cite{yang2014deep} & 25.6M & 56.18 & 2.3x \\
 Adaptive FastFood 32 \cite{yang2014deep}& 22.5M & 57.39 & 2.7x  \\
 Adaptive FastFood 16 \cite{yang2014deep}& 16.4M & 57.1 & 3.7x  \\ 
 ACDC \cite{moczulski2015acdc} & 11.9M & 56.73 & 5x \\
 Collins \& Kohli \cite{DBLP:journals/corr/CollinsK14}  & 8.5M (17M) & 55.60 &7x \\
 Han \etal \cite{han2015learning} & 6.7M (13.4M)   & 57.2 & 9x \\
 
 \hline
 \textbf{Proposed Method} & \textbf{5.9M} (11.8M) & \textbf{56.96} & \textbf{10.3x} \\
\hline
\end{tabular}

   \caption{Comparison of compression performance on AlexNet architecture}
    \label{table:alexnetcompare}
\end{table*}

\begin{table*}[!th]
\centering
\begin{tabular}{ |c|c|c |c| }
 \hline
 \textbf{Method} & \textbf{Params (Params + Indices)} & \textbf{Top-1 Accuracy(\%)} & \textbf{Compression Rate}\\
 \hline
 \hline
 Baseline  & 138M   &  68.97 & - \\
 
 Han \etal \cite{han2015learning} & 10.3M (20.6M)   & 68.66 & 13x \\

 \textbf{Proposed Method} & \textbf{9.8M} (19.6M) & \textbf{69.04} & \textbf{14x} \\
\hline
\end{tabular}

   \caption{Comparison of compression performance on VGG-16 architecture}
    \label{table:vggnetcompare}
\end{table*}

We now study the effects of using different initializations for the gate parameters. We initialize all gate parameters of a layer with the same constant value. We also tried stochastic initialization for these gate parameters (Eg. from a Gaussian distribution), but we found no particular advantage in doing so. As shown in Figure \ref{fig:analyzeinit}, both methods seem robust to varying initializations, with the thresholding method consistently giving higher sparsities. This robustness to initialization is advantageous to our method, as we no longer need to worry about finding good initial values for them.



\subsection{Compression Performance}
We test compression performance on three different network architectures - LeNet-5, AlexNet \cite{krizhevsky2012imagenet} and VGG-16. 

For LeNet-5, we simply sparsify each layer. As shown in Table \ref{table:lenetdetailed}, we are able to remove about 96\% of LeNet's parameters and only suffer a negligible loss in accuracy. Table \ref{table:lenetcompare} shows that we obtain state-of-the-art results on LeNet-5 compression. For Proposed Method - 1, we used $(\lambda_1 , \lambda_2) = (0.001, 0.05)$, while for Proposed Method-2, we used $(\lambda_1 , \lambda_2) = (0.01, 0.1)$. These choices were made using a validation set.

Note that our method converts a dense matrix to a sparse matrix, so the total number of parameters that need to be stored on disk includes the indices of the parameters. As a result, we report the parameter count along with indices. This is similar to what has been done in \cite{DBLP:journals/corr/CollinsK14}. However, for ASIC implementations, one need not store indices as they can be built into the circuit structure. 

For AlexNet and VGG-16, instead of training from scratch, we fine-tune the network from pre-trained weights. For such pre-trained weights, we found it be useful to pre-initialize the gate variables so that we do not lose accuracy while starting to fine-tune. Specifically, we ensure that the gate variables corresponding to the top-$k \%$ weights in the $W$ matrix are one, while the rest are zeros. We use this pre-initialization instead of the constant initialization described previously. 

To help pruning performance, we pre-initialize fully connected gates with very large sparsity ($95 \%$) and convolutional layers with very little sparsity. This means that $95 \%$ of $g^s$ parameters are zero, and rest are one. For $g^s = 1$, the underlying gate values were $g = 1$ and for $g^s = 0$, we used $g = 0.49$. This is to ensure good accuracy by preserving important weights while having large sparsity ratios. The resulting network ended up with a negligible amount of sparsity for convolutional layers and high sparsity for fully connected layers. For VGG-16, we pre-initialize the final two convolutional layers as well with $88 \%$ sparsity. 

We run fine-tuning on AlexNet for 30k iterations ($\sim 18$ hours), and VGG for 40k ($\sim 24$ hours) iterations before stopping training based on the combination of compression ratio and validation accuracy. This is in contrast with \cite{han2015learning}, who take about 173 hours to fine-tune AlexNet. The original AlexNet took 75 hours to train. All wall clock numbers are reported by training on a NVIDIA Titan X GPU. As shown in Table \ref{table:alexnetcompare} and Table \ref{table:vggnetcompare}, we obtain favourable results when compared to the other network compression / sparsification methods.

\section{Conclusion}
We have introduced a novel method to learn neural networks with sparse connections. This can be interpreted as learning weights and performing pruning simultaneously. By introducing a learning-based approach to pruning weights, we are able to obtain the optimal level of sparsity. This enables us to achieve state-of-the-art results on compression of deep neural networks.

\bibliographystyle{aaai}

\bibliography{sparse_nets}

\end{document}